\documentclass[a4paper,fleqn]{cas-sc}

\usepackage[numbers]{natbib}

\def\tsc#1{\csdef{#1}{\textsc{\lowercase{#1}}\xspace}}
\tsc{WGM}
\tsc{QE}
\tsc{EP}
\tsc{PMS}
\tsc{BEC}
\tsc{DE}

\usepackage{epigraph}
\usepackage{longtable}
\usepackage{setspace}

\usepackage{textcomp}
\usepackage{float}

\usepackage{multirow}

\usepackage{color}

\definecolor{WordGreen}{RGB}{100, 136, 40}
\definecolor{WordDarkGrey}{RGB}{82, 82, 82}
\definecolor{WordRed}{RGB}{192, 80, 77}
\definecolor{WordBlue}{RGB}{0, 122, 192}

\definecolor{WordLightBlue}{RGB}{218, 238, 243}
\definecolor{WordLightGreen}{RGB}{234, 241, 221}

\definecolor{WordFillGreen}{RGB}{194, 214, 155}
\definecolor{WordFillRed}{RGB}{252, 214, 182}
\definecolor{WordFillGray}{RGB}{217, 217, 217}

\UseRawInputEncoding
\usepackage{times}
\usepackage{epsfig}
\usepackage{amssymb}
\usepackage[ruled,vlined,linesnumbered, longend]{algorithm2e}
\usepackage{geometry}
\usepackage{comment}
\usepackage{array, makecell} 
\usepackage{lettrine}
\usepackage{url} 
\usepackage{lscape}
\usepackage{cuted}

\usepackage{algpseudocode}  
\usepackage{tabularx}
\usepackage{colortbl}
\usepackage[normalem]{ulem}
\usepackage{hhline}
\usepackage{vcell}
\usepackage{longtable}
\usepackage{vcell}
\usepackage{rotating}
\usepackage{pdflscape}
\usepackage{sidecap}
\usepackage{neuralnetwork}
\usepackage[export]{adjustbox} 
\usepackage[accsupp]{axessibility} 
\usepackage{amsfonts,bm}
\usepackage{physics}
\usepackage{wrapfig}
\usepackage{caption}
\usepackage{color,soul}
\usepackage[frozencache,cachedir=minted-cache]{minted}
\usepackage{mathtools}
\usepackage{amsthm}

\usepackage{acronym}
\acrodef{FCN}[FCN]{Fully Convolutional Network}
\acrodef{GAME}[GAME]{Grid Average Mean Absolute Error}
\acrodef{DL}[DL]{Deep Learning}
\acrodef{DNN}[DNN]{Deep Neural Network}
\acrodef{ML}[ML]{Machine Learning}
\acrodef{CV}[CV]{Computer Vision}
\acrodef{AI}[AI]{Artificial Intelligence}
\acrodef{CNN}[CNN]{Convolutional Neural Network}
\acrodef{RNN}[RNN]{Recurrent Neural Network}
\acrodef{GAN}[GAN]{Generative Adversarial Network}
\acrodef{JCU}[JCU]{James Cook University}
\acrodef{MAE}[MAE]{Mean Average Error}
\acrodef{MAP}[mAP]{Mean Average Precision}
\acrodef{CA}[CA]{Classification Accuracy}
\acrodef{LCFCN}[LCFCN]{Localization-based Counting loss Fully Convolutional Network}
\acrodef{IoT}[IoT]{Internet of Things}
\acrodef{MLP}[MLP]{Multi-Layer Perceptrons}

\def\btheta{{\bm\theta}}

\DeclareMathOperator*{\argmax}{arg\,max}

\def\lL{\mathcal{L}}
\def\Re{\mathbb{R}}
\usepackage{xspace}

\def\dd{\mathbf{d}}
\usepackage[capitalize]{cleveref}
\crefname{section}{Sec.}{Secs.}
\Crefname{section}{Section}{Sections}
\Crefname{table}{Table}{Tables}
\crefname{table}{Table}{Table}
\usepackage{booktabs}
\usepackage{arydshln}
\usepackage{listings}
\definecolor{codegreen}{rgb}{0,0.6,0}
\definecolor{codegray}{rgb}{0.5,0.5,0.5}
\definecolor{codepurple}{rgb}{0.58,0,0.82}
\definecolor{backcolour}{rgb}{0.95,0.95,0.92}
\definecolor{cleacolorr}{rgb}{1,1,1}

\lstdefinestyle{mystyle}{
    backgroundcolor=\color{cleacolorr},   
    commentstyle=\color{codegreen},
    keywordstyle=\color{magenta},
    numberstyle=\tiny\color{codegray},
    stringstyle=\color{codepurple},
    basicstyle=\ttfamily\footnotesize,
    breakatwhitespace=false,         
    breaklines=true,                 
    captionpos=b,                    
    keepspaces=true,                 
    numbers=left,                    
    numbersep=5pt,                  
    showspaces=false,                
    showstringspaces=false,
    showtabs=false,                  
    tabsize=2
}

\usepackage[most]{tcolorbox}
\newtcolorbox[auto counter]{pabox}[2][]{%
colback=blue!5!white,colframe=blue!75!black,fonttitle=\bfseries,
title=Box~\thetcbcounter: #2,#1}

\usepackage{tikz}

\usepackage{graphicx}
\graphicspath{{./Graphics/}}
\DeclareGraphicsExtensions{.pdf,.jpeg,.png,.jpg}

\usepackage{amsmath}

\usepackage{array}

\usepackage{url}

\begin{document}

\let\WriteBookmarks\relax
\def\floatpagepagefraction{1}
\def\textpagefraction{.001}

\shorttitle{Depth and Surface Normals Estimation}

\shortauthors{Saleh et~al.}

\title [mode = title]{
A Practical Approach to Underwater Depth and Surface Normals Estimation
}   




%



\author[1]{Alzayat Saleh}[orcid=0000-0001-6973-019X]
\cormark[1]
\ead{alzayat.saleh@my.jcu.edu.au}
\credit{Conceptualisation, Data Curation, Data Analysis, Software Development, DL Algorithm Design, Visualization, Writing original draft}

\author[2]{Melanie Olsen}
\ead{m.olsen@aims.gov.au}
\credit{Conceptualisation, Supervision, Reviewing/editing the draft}

\author[1]{Bouchra Senadji}
\ead{bouchra.senadji@jcu.edu.au}
\credit{Conceptualisation, Supervision, Reviewing/editing the draft}

\author[1]{Mostafa Rahimi Azghadi}[                        orcid=0000-0001-7975-3985]
\ead{Mostafa.Rahimiazghadi@jcu.edu.au}

\credit{Conceptualisation, Supervision, Reviewing/editing the draft}





\affiliation[1]{organization={College of Science and Engineering, James Cook University},
    city={Townsville city},
    state={Queensland},
    country={Australia}}



\affiliation[2]{organization={The Australian Institute of Marine Science},
    city={Townsville city},
    state={Queensland},
    country={Australia}}




\cortext[cor1]{Corresponding author}




\begin{abstract}
Monocular Depth and Surface Normals Estimation (MDSNE) is crucial for tasks such as three-dimensional (3D) reconstruction, autonomous navigation, and underwater exploration. Current methods rely either on discriminative models, which struggle with transparent or reflective surfaces, or generative models, which, while accurate, are computationally expensive. This paper presents a novel deep learning model for Monocular Depth and Surface Normals Estimation (MDSNE), specifically tailored for underwater environments, using a hybrid architecture that integrates Convolutional Neural Networks (CNNs) with Transformers, leveraging the strengths of both approaches.  
Training effective MDSNE models is often hampered by noisy real-world datasets and the limited generalization of synthetic datasets. To address this, we generate pseudo-labeled real data using multiple pre-trained MDSNE models. To ensure the quality of this data, we propose the Depth Normal Evaluation and Selection Algorithm (DNESA), which evaluates and selects the most reliable pseudo-labeled samples using domain-specific metrics. A lightweight student model is then trained on this curated dataset.  
Our model reduces parameters by 90\% and training costs by 80\%, allowing real-time three-dimensional (3D) perception on resource-constrained devices. Key contributions include: a novel and efficient MDSNE model, the DNESA algorithm, a domain-specific data pipeline, and a focus on real-time performance and scalability. Designed for real-world underwater applications, our model facilitates low-cost deployments in underwater robots and autonomous vehicles, bridging the gap between research and practical implementation.
\end{abstract}





\maketitle


 \begin{figure}[h]
    \centering
    \includegraphics[width=0.98\linewidth]{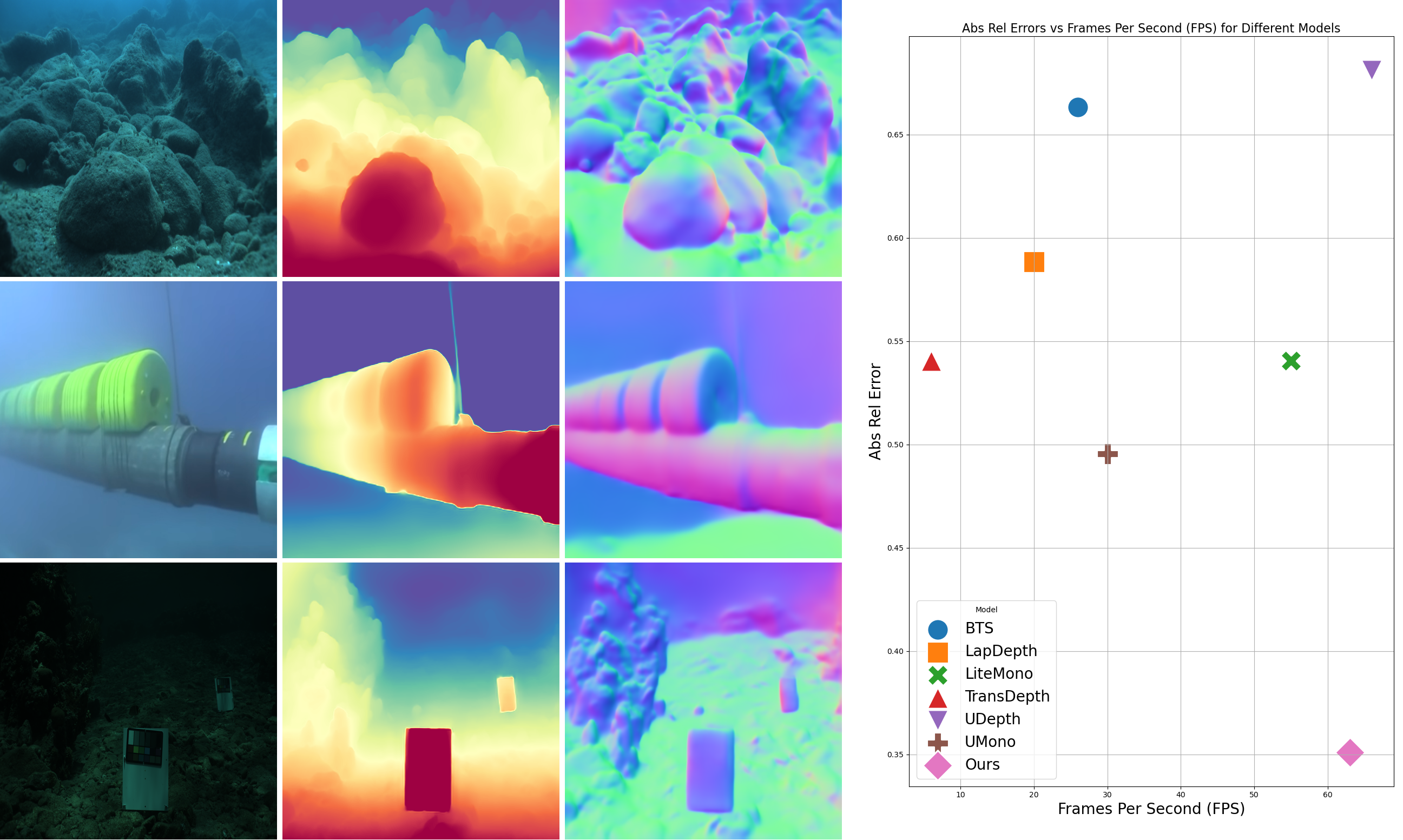}
    \caption{
\textbf{Integrating the strengths of both discriminative and generative approaches for geometry estimation.} Left: Depth and normal
predictions of our method on in-the-wild images. Right: Abs Rel Errors vs Frames Per Second (FPS) for Different Models
}
    \label{l_fig_1}
\end{figure}


\section{Introduction}

Monocular Depth and Surface Normals Estimation (MDSNE) plays a critical role in various applications, such as 3D reconstruction, autonomous navigation, and underwater exploration. Existing MDSNE approaches generally fall into two categories: discriminative models and generative models. Discriminative models, like Depth Anything \cite{yang2024depth}, excel at producing robust depth predictions efficiently, even in complex scenes, but struggle with transparent or reflective surfaces. However, generative models, like Marigold \cite{ke2024repurposing}, capture intricate object details, providing more refined depth maps, but are less practical in real-time or resource-constrained situations. 

This performance gap between discriminative and generative models highlights the need for an approach that combines the strengths of both, offering both efficiency and detailed precision.  Furthermore, training accurate and robust MDSNE models often relies on large-scale datasets. However, real-world datasets often contain noisy labels, particularly for transparent or reflective objects, and lack fine-grained depth and surface normals annotations, leading to suboptimal predictions. Although synthetic datasets offer precise labels and consistent annotations, they suffer from domain gaps when applied to real-world scenarios. 

This paper introduces a novel deep learning model for MDSNE that addresses these challenges by merging the strengths of discriminative and generative approaches within a lightweight architecture, specifically designed for the complex underwater environment.  We achieve this by first generating high-quality pseudo-labelled depth and surface normal data from unlabelled real underwater images using multiple state-of-the-art pre-trained models. However, generating high-quality pseudo-labels  is not straightforward or simple, as it requires careful selection and evaluation to ensure the accuracy of the generated labels.  To address this, we introduce the Depth Normal Evaluation and Selection Algorithm (DNESA) to evaluate the quality of the generated pseudo-depth and surface normals.  DNESA helps select the most reliable pseudo-labeled data, ensuring that our training process is not compromised by inaccuracies in the labels.  We then train a lightweight student model on this carefully curated pseudo-labeled dataset, combining a hybrid architecture of Convolutional Neural Networks (CNNs) and Transformers to capture both local and global image features effectively.  This approach allows us to leverage the strengths of both synthetic and real-world data, leading to a model that is accurate and computationally efficient. 

The paper addresses key limitations in Monocular Depth and Surface Normals Estimation (MDSNE) for underwater environments, with five main contributions:

\begin{itemize}
    \item \textbf{Novel MDSNE Model:} A hybrid model combining CNNs for local features and Transformers for global context, reducing parameters by 90\% and training costs by 80\% compared to Marigold \cite{ke2024repurposing}, making it suitable for low-power devices.
    
    \item \textbf{Depth Normal Evaluation and Selection Algorithm (DNESA):} An algorithm to assess and select reliable pseudo-labeled data based on underwater-specific challenges like light refraction and turbidity.

    \item \textbf{Domain-Specific Data Pipeline:} A pipeline tailored for underwater visuals, addressing low light, turbidity, and colour variation to improve model performance.

    \item \textbf{Real-Time 3D Perception:} The model achieves real-time inference (approximately 15.87ms per image), enabling use in tasks like underwater robot navigation.

    \item \textbf{Scalability and Practicality:} Prioritising efficiency and accessibility, the model supports scalability for broader use in real-world underwater applications.
\end{itemize}

The rest of this paper is organised as follows: Section \ref{sec:related_work} reviews related work on underwater depth estimation and surface normals, focussing on state-of-the-art approaches and their limitations in resource-constrained environments.  
Section \ref{sec:methodology} details our proposed methodology, including the hybrid CNN-Transformer architecture and the domain-specific data pipeline.  Section \ref{sec:experiments} presents the experimental setup and evaluates the performance of our model compared to existing methods, highlighting its efficiency and precision.  
Section \ref{sec:discussion} provides an in-depth discussion of the impact of the model on marine science tasks and outlines potential applications.  Section \ref{sec:conclusion} concludes the paper, summarising the contributions and suggesting directions for future research.

 
\section{Related Work} \label{sec:related_work}

Research in Monocular Depth and Surface Normals Estimation (MDSNE) has significantly advanced, driven by the increasing demand for 3D perception in fields like robotics, autonomous navigation, and underwater exploration. This section provides an overview of key contributions in MDSNE, categorising them into discriminative and generative models, and discusses the challenges posed by different data sources for training these models.

\vspace{-0.7em} \paragraph{Discriminative Models,}
 have gained prominence in MDSNE due to their ability to produce robust depth predictions efficiently, even in complex scenes. These models typically rely on Convolutional Neural Networks (CNNs) to learn a direct mapping from input images to depth maps. A prime example is Depth Anything \cite{yang2024depth}, a model known for its efficiency and robustness in handling cluttered scenes. However, discriminative models often struggle with transparent or reflective surfaces, as they primarily focus on local features and struggle to capture global context \cite{costanzino2023learning}.

\vspace{-0.7em} \paragraph{Generative Models,}
 such as Marigold \cite{ke2024repurposing}, address some limitations of discriminative models by learning to generate depth and surface normals from scene representations. These models excel at capturing fine-grained details and handling challenging surface properties like transparency and reflectivity. However, they typically come with a higher computational cost, making them less practical for real-time applications or resource-constrained environments \cite{yin2023metric3d}.

\vspace{-0.7em} \paragraph{Synthetic vs. Real Data for MDSNE:}

Training effective MDSNE models heavily relies on the availability of large-scale datasets. These datasets can be categorised into synthetic and real data, each presenting unique advantages and challenges.
1) \textit{Synthetic Datasets:} These datasets provide precise depth labels and surface normals due to the controlled environment in which they are generated. However, their main drawback lies in the domain gap between synthetic and real-world scenarios, which often leads to decreased performance when applied to real-world settings \cite{atlantis}.
2) \textit{Real Datasets:} These datasets offer greater realism and scene diversity but suffer from noisy labels, particularly for transparent or reflective objects. They often lack the fine-grained annotations required for accurate depth and surface normal estimations \cite{yin2023metric3d}.

\vspace{-0.7em} 
\paragraph{Bridging the Gap:}

Our method leverages the strengths of both discriminative and generative models, addressing their individual limitations through a lightweight, hybrid CNN-Transformer architecture. This approach balances accuracy and computational efficiency, making it suitable for real-time deployment on resource-constrained devices, unlike more resource-intensive diffusion-based models \cite{zavadski2024primedepth}.

To overcome the challenge of noisy labels in real-world data, we introduce a novel pseudo-labelling technique. By utilising state-of-the-art models, we generate high-quality pseudo-labelled data from unlabeled real underwater images, combining the efficiency of discriminative models with the detail-oriented capabilities of generative models. This approach ensures robust generalisation across diverse underwater environments \cite{USOD10K, atlantis, Sea-Thru, SQUID, DRUVA, DeepFish}.


\begin{figure}[t]
    \centering
    \includegraphics[width=0.98\linewidth]{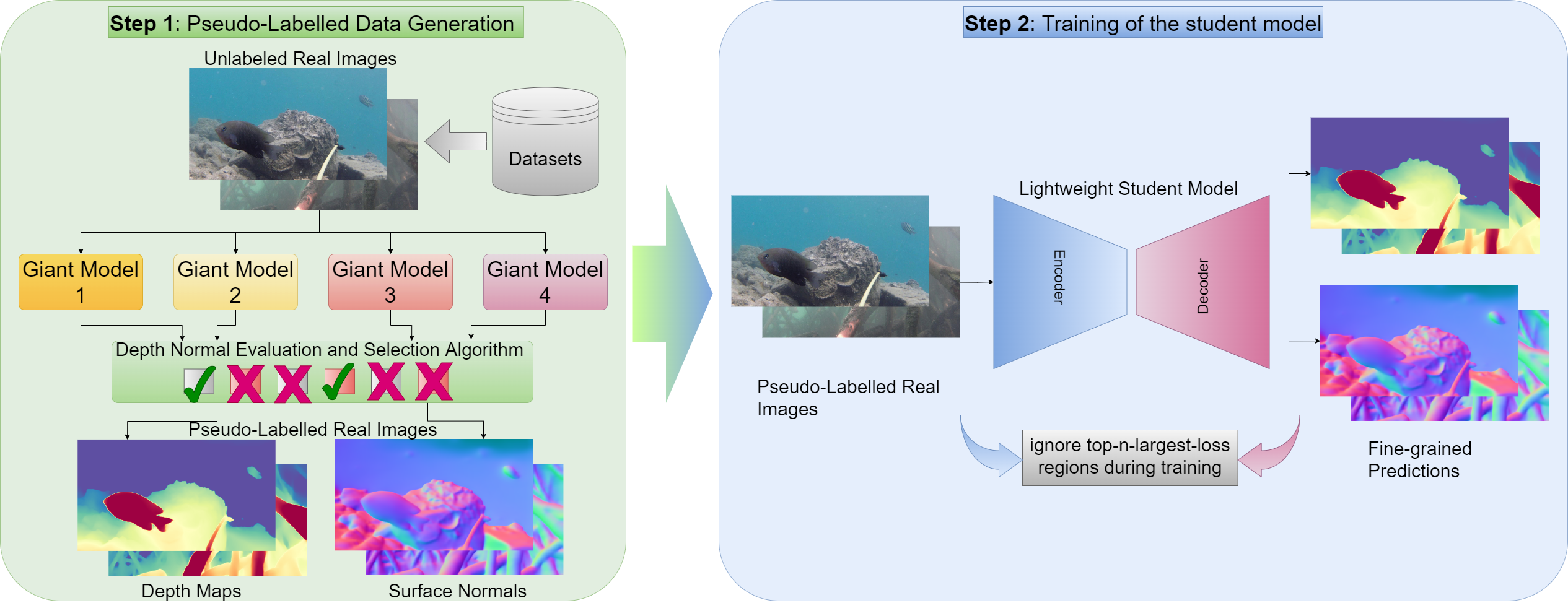}
    \caption{\textbf{Our Framework}: Two-step process for generating pseudo-labeled data and training a student model. \textbf{Step 1} involves generating pseudo-labeled data by inputting real images into multiple Giant Models to produce depth maps and surface normals, followed by evaluation and selection of the best outputs. \textbf{Step 2} involves training a Lightweight Student Model using the pseudo-labeled real images, highlighting regions to ignore during training due to high-error loss, and resulting in fine-grained predictions.}
    \label{l_fig_4}
\end{figure}

\section{Method} \label{sec:methodology}

 
\subsection{Learning from the Giants}
This section presents our analysis of state-of-the-art models, which motivated us to develop our approach.

\vspace{-0.7em}
\paragraph{Key Findings:}
Through this research, we identified several critical insights into the development of MDSNE models:
\begin{enumerate}
    \item Fine details in depth maps can be achieved using discriminative models when trained on synthetic images, which offer more precise control over scene content.
    \item Despite the superiority of synthetic images for detail-oriented tasks, real images are still widely used due to the challenges of fully addressing the limitations of synthetic data.
    \item The limitations of synthetic data, such as distribution shifts and scene diversity, can be mitigated by creating large-scale pseudo-labeled real datasets, improving the robustness of the models.
\end{enumerate}

 \begin{figure}[t]
    \centering
    \includegraphics[width=0.98\linewidth]{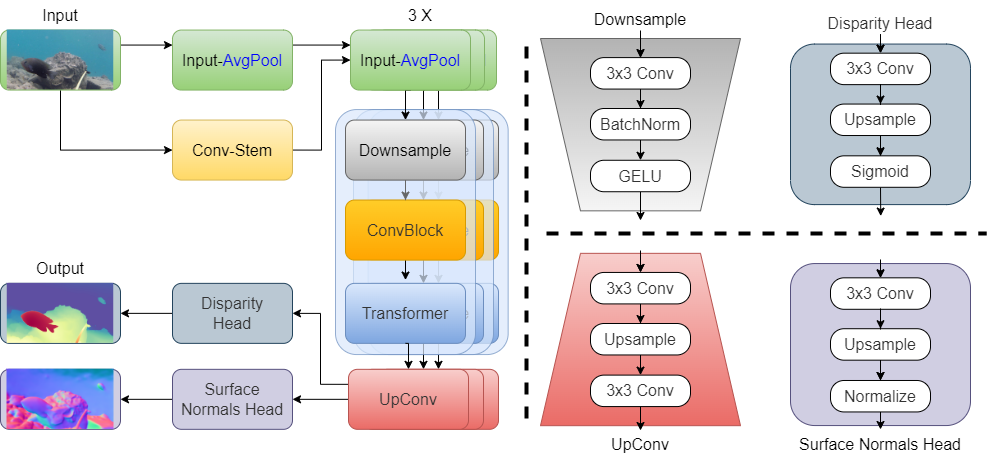}
    \caption{\textbf{Schematic of our network architecture for depth and surface normals estimation}. This diagram delineates the data flow through various components, including the Conv-Stem, ConvBlock, and Transformer, culminating in two parallel branches. The upper branch is dedicated to depth estimation, incorporating upsampling, convolutional layers, and a Disparity Head. The lower branch is focused on surface normals estimation, featuring upsampling, convolutional layers, and a Surface Normals Head. This architecture underscores the integration of convolutional and transformer blocks to enhance the accuracy of depth and surface normals estimation.}
    \label{l_fig_2}
\end{figure}

\paragraph{Challenges with Real Labeled Data:}
Real-world datasets often introduce significant challenges in MDSNE due to inherent noise in labels, particularly with objects that possess transparent or reflective properties. Furthermore, real datasets frequently lack fine-grained depth and surface normals annotations for smaller or thinner objects, leading to suboptimal predictions in such scenarios. This makes it difficult to achieve the same level of precision and accuracy that can be derived from synthetic data, where labels are precise and controlled.

\paragraph{Advantages and Limitations of Synthetic Data:}
Synthetic data offer clear advantages in providing precise depth labels and consistent annotations, but they are not without limitations. The primary challenge with synthetic data lies in the distribution shift between synthetic and real-world scenarios, which can hinder model performance when applied to real environments. In addition, synthetic datasets often lack the diversity and complexity of real-world scenes, making it difficult to generalise effectively between tasks. Synthetic-to-real transfer, therefore, remains a significant challenge in the deployment of MDSNE models.

\paragraph{Our Approach:}
According to all the above analyses, our proposed approach is designed with a focus on both precision and computational efficiency, particularly in the context of real-time applications. The general framework consists of two main steps: (1) generating precise pseudo-depth and surface normals labels from large-scale, unlabelled real images, and (2) training a final student model that balances accuracy with computational efficiency for deployment on edge devices, ensuring robust generalisation across diverse environments,  as illustrated in Figure \cref{l_fig_4}.

 \begin{figure}[t]
    \centering
    \includegraphics[width=0.70\linewidth]{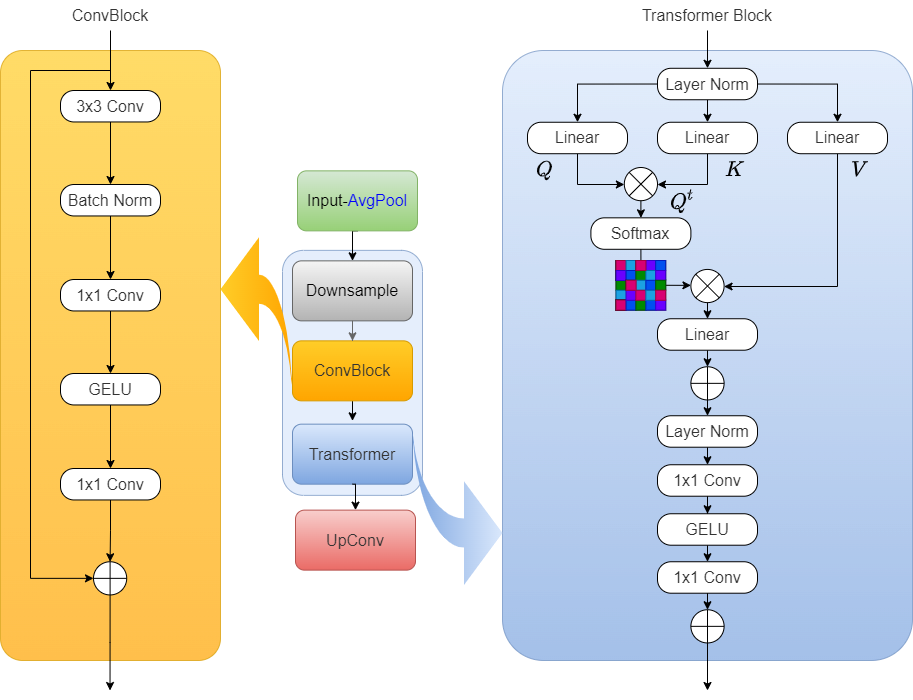}
    \caption{Schematic  diagram illustrates the components and flow of data through ConvBlock, Transformer Block, and associated processes.}
    \label{l_fig_3}
\end{figure}

\subsection{Step 1: Pseudo-Labelled Data Generation}

In the first step, our objective is to address the limitations of real-world labelled datasets, which often suffer from noise, particularly with reflective, transparent, or fine-grained objects. To overcome this, we utilise a combination of state-of-the-art discriminative and generative models to create high-quality pseudo-labeled data. These models are pre-trained on synthetic and real datasets, which provide precise depth and surface normal annotations. These models are DepthAnything V2 \cite{yang2024depth}, ZoeDepth  \cite{bhat2023zoedepth}, Metric3D -V2 \cite{yin2023metric3d}, and Marigold \cite{ke2024repurposing}.
Using these pre-trained models, we infer pseudo depth and surface normals on large-scale, unlabeled real images. This process helps bridge the gap between synthetic and real data, capturing the complexities and variations present in real-world scenes while maintaining the accuracy associated with synthetic annotations. The generated pseudo-labeled data thus serves as a more reliable and diverse dataset for training.

While these state-of-the-art pre-trained models demonstrate exceptional performance and robustness in general depth and surface normals estimation tasks, they may encounter challenges when applied to underwater environments. The unique visual characteristics of underwater scenes, such as low-light conditions, water turbidity, and variations in colour spectra, can introduce a domain gap between the training data and the target underwater images. This domain gap can lead to inaccuracies in depth maps estimated for out-of-distribution underwater scenarios.

To address this challenge, we propose the Depth Normal Evaluation and Selection Algorithm (DNESA). This algorithm is designed to evaluate the quality of the generated pseudo depth and surface normals in underwater environments and select the most accurate samples for training. DNESA operates by comparing the pseudo-labeled depth and normals against domain-specific metrics, such as contrast variation, edge sharpness, and noise levels, which are more representative of underwater conditions.
Once the pseudo-labeled data have been evaluated and filtered through DNESA, the remaining high-quality samples are added to the training pipeline. This approach ensures that the final training dataset is not only diverse but also better suited to underwater domains, reducing the impact of domain gaps and improving overall model performance when deployed in real-world underwater tasks.

\begin{algorithm}[ht]
\footnotesize
\SetAlgoNlRelativeSize{0.1} 
\caption{Depth Normal Evaluation and Selection Algorithm (DNESA)}
\label{code1}
\LinesNumbered
\KwIn{RGB folder path, Model folders paths, Output folder path}
\KwOut{Best depth map and surface normals for each RGB image are copied to the output folder}
\SetKwFunction{FMain}{Main}
\SetKwProg{Fn}{Function}{:}{}
\Fn{\FMain{rgbFolder, modelFolders, outputFolder}}{
    \ForEach{rgbFile in rgbFolder}{
        rgbImage $\gets$ LoadImage(rgbFile)\;
        depthResults $\gets$ \{\}\;
        normalResults $\gets$ \{\}\;
        depthPaths $\gets$ \{\}\;
        normalPaths $\gets$ \{\}\;
        \ForEach{modelFolder in modelFolders}{
            depthFile $\gets$ FindCorrespondingDepthFile(rgbFile, modelFolder)\;
            normalFile $\gets$ FindCorrespondingNormalFile(rgbFile, modelFolder)\;
            \If{depthFile $\neq$ null \textbf{and} normalFile $\neq$ null}{
                depthMap $\gets$ LoadDepthMap(depthFile)\;
                normalMap $\gets$ LoadNormalMap(normalFile)\;
                depthResult $\gets$ EvaluateDepthMap(depthMap, rgbImage)\;
                normalResult $\gets$ EvaluateNormalMap(normalMap, rgbImage)\;
                depthResults.append(depthResult)\;
                normalResults.append(normalResult)\;
                depthPaths.append(depthFile)\;
                normalPaths.append(normalFile)\;
            }
        }
        \If{depthResults $\neq$ \{\} \textbf{and} normalResults $\neq$ \{\}}{
            bestIndex $\gets$ FindBestDepthNormalPair(depthResults, normalResults)\;
            bestDepthPath $\gets$ depthPaths[bestIndex]\;
            bestNormalPath $\gets$ normalPaths[bestIndex]\;
            CopyFile(bestDepthPath, outputFolder)\;
            CopyFile(bestNormalPath, outputFolder)\;
        }
    }
}
\Fn{EvaluateDepthMap{depthMap, rgbImage}}{
    depthMap $\gets$ NormalizeDepthMap(depthMap)\;
    edgeConsistency $\gets$ ComputeEdgeConsistency(depthMap, rgbImage)\;
    localVariance $\gets$ ComputeLocalDepthVariance(depthMap)\;
    complexity $\gets$ ComputeDepthComplexity(depthMap)\;
    sharpness $\gets$ ComputeSharpnessMetric(depthMap)\;
    \Return \{edgeConsistency, localVariance, complexity, sharpness\}\;
}
\Fn{EvaluateNormalMap{normalMap, rgbImage}}{
    normalMap $\gets$ NormalizeNormalMap(normalMap)\;
    edgeConsistency $\gets$ ComputeEdgeConsistency(normalMap, rgbImage)\;
    orientationVariance $\gets$ ComputeNormalorientationVariance(normalMap)\;
    sharpness $\gets$ ComputeSharpnessMetric(normalMap)\;
    \Return \{edgeConsistency, orientationVariance, sharpness\}\;
}
\Fn{FindBestDepthNormalPair{depthResults, normalResults}}{
    depthWeights $\gets$ \{edgeConsistency: 0.3, localVariance: -0.2, complexity: 0.2, sharpness: 0.3\}\;
    normalWeights $\gets$ \{edgeConsistency: 0.4, orientationVariance: 0.4, sharpness: 0.2\}\;
    combinedScores $\gets$ \{\}\;
    \ForEach{(depthResult, normalResult) in zip(depthResults, normalResults)}{
        depthScore $\gets$ 0\;
        normalScore $\gets$ 0\;
        \ForEach{metric in depthWeights}{
            depthScore $\gets$ depthScore + depthResult[metric] $\times$ depthWeights[metric]\;
        }
        \ForEach{metric in normalWeights}{
            normalScore $\gets$ normalScore + normalResult[metric] $\times$ normalWeights[metric]\;
        }
        combinedScore $\gets$ depthScore + normalScore\;
        combinedScores.append(combinedScore)\;
    }
    \Return $\argmax(combinedScores)$\;
}
\end{algorithm}

\paragraph{\textbf{Depth Normal Evaluation and Selection Algorithm (DNESA):}}

The Depth Normal Evaluation and Selection Algorithm (DNESA), as outlined in Algorithm~\ref{code1}, addresses the challenge of noisy labels commonly present in real-world underwater datasets. DNESA systematically evaluates the quality of pseudo-depth maps and surface normals derived from unlabelled underwater images, leveraging multiple pre-trained models. The algorithm assesses these outputs using a set of predefined metrics and selects the most reliable data to train a lightweight student model.
DNESA operates in two primary stages. The first stage involves evaluating the pseudo-labeled data by computing several critical metrics:

\begin{itemize}
    \item \textbf{Edge Consistency:} This metric quantifies the alignment between the edges of the depth map or surface normal map and the corresponding edges in the RGB image. Given an RGB image $\mathbf{I}_{\text{RGB}}$ and a depth map $\mathbf{D}$, edge consistency is evaluated as:
    $
    E_c = \frac{|\nabla \mathbf{D}| \cap |\nabla \mathbf{I}_{\text{RGB}}|}{|\nabla \mathbf{I}_{\text{RGB}}|}
    $
    where $\nabla$ denotes the gradient operation.
    
    \item \textbf{Local Variance:} Measures the variability of depth values or surface normal orientations within local neighbourhoods. For a depth map $\mathbf{D}$, local variance $V$ within a window $W$ is computed as:
    $
    V = \frac{1}{|W|} \sum_{i \in W} \left( D_i - \mu_W \right)^2,
    $
    where $\mu_W$ is the mean depth value within window $W$.

    \item \textbf{Complexity:} Complexity quantifies the structural intricacy of the depth map or surface normal map by assessing the spatial frequency content, expressed as:
    $
    C = \frac{1}{N} \sum_{i=1}^{N} |\nabla \mathbf{D}_i|,
    $
    where $N$ is the number of pixels in the map.

    \item \textbf{Sharpness:} Sharpness evaluates the clarity of the edges in the depth or normal maps, defined by the maximum gradient magnitude across the map:
    $
    S = \max\left( |\nabla \mathbf{D}| \right).
    $

    \item \textbf{Orientation Variance:} For surface normals, orientation variance assesses the alignment of predicted normal orientations. Given a set of normal vectors $\mathbf{n}_i$ and their mean $\bar{\mathbf{n}}$, this metric is expressed as:
    $
    V_o = \frac{1}{N} \sum_{i=1}^{N} \left( 1 - \mathbf{n}_i \cdot \bar{\mathbf{n}} \right),
    $
    where $\cdot$ denotes the dot product.
\end{itemize}

In the second stage, DNESA selects the most reliable depth and surface normal maps by ranking them based on a weighted scoring system. For depth maps, the metrics are weighted as follows: edge consistency ($w_1 = 0.3$), sharpness ($w_2 = 0.3$), local variance ($w_3 = -0.2$), and complexity ($w_4 = 0.2$). The final score for each depth map $\mathbf{D}$ is computed as:
$
S_{\mathbf{D}} = w_1 E_c + w_2 S - w_3 V + w_4 C.
$
For surface normals, edge consistency ($w_1 = 0.4$) and orientation variance ($w_2 = 0.4$) are prioritised, with sharpness ($w_3 = 0.2$) contributing to a lesser extent. The score for surface normals $\mathbf{N}$ is:
$
S_{\mathbf{N}} = w_1 E_c + w_2 V_o + w_3 S.
$

The \texttt{FindBestDepthNormalPair} function, as detailed in Algorithm~\ref{code1}, selects the depth-normal pair that achieves the highest combined score. This process ensures that the student model is trained on the highest-quality pseudo-labeled data, which is crucial for achieving reliable 3D perception in underwater environments.


\subsection{Step 2: Training of the Student Model}

Following the generation of a robust pseudo-labelled real dataset, the next step involves training a lightweight student model. This model is engineered for computational efficiency while preserving the accuracy achieved by the more complex teacher models used in the pseudo-labelling phase. The student model adopts a hybrid architecture that combines Convolutional Neural Networks (CNNs) with Transformers, as depicted in Figure \ref{l_fig_2}.

The student model architecture comprises two key components: the encoder and the decoder. Each component is carefully designed to achieve an optimal balance between performance and computational efficiency, making it particularly suitable for deployment on resource-constrained edge devices. The encoder, leveraging CNN layers, efficiently extracts feature maps, while the inclusion of Transformer layers enhances the model's ability to capture long-range dependencies and global context.

The architecture is also scalable, allowing for adjustments in channel numbers, convolutional blocks, and dilation rates. This scalability ensures the model's adaptability to varying computational constraints and resource availability. The flexibility in design enables the model to maintain high performance across a wide range of devices, from high-powered servers to low-power embedded systems. This makes the student model versatile, capable of delivering reliable results even when deployed in environments with limited computational resources.

\subsubsection{Encoder Architecture Design}

The encoder, as outlined in Algorithm~\ref{alg:Encoder}, is tasked with extracting hierarchical feature representations from the input image, denoted as $\mathbf{x} \in \mathbb{R}^{H \times W \times C}$. The first step involves normalising the input image using the mean $\mu$ and standard deviation $\sigma$ across the dataset: 
\begin{equation}
    \mathbf{x}_{norm} = \frac{\mathbf{x} - \mu}{\sigma}.
\end{equation}

Following normalisation, a series of downsampling operations is applied to progressively reduce the spatial resolution while retaining essential visual information. This multi-scale downsampling is achieved using average pooling over different scales, such that for each scale $i$, the pooled representation is denoted as:
\begin{equation}
    \mathbf{x}_{down}^i = \text{AvgPool}(\mathbf{x}, i), \quad i \in [1, 5].
\end{equation}

The core of the encoder is organised into three distinct stages, each comprising a combination of convolutional and transformer blocks, as depicted in Figure~\ref{l_fig_3}. The convolutional layers employ dilated convolutions, represented as $\text{Conv}_{\text{dilation}}$, to expand the receptive field without a significant increase in computational cost. Specifically, for a convolution operation with kernel size $k$ and dilation factor $d$, the receptive field is increased by a factor of $d$:
\begin{equation}
    \text{Receptive Field} = (k - 1) \cdot d + 1.
\end{equation}

In parallel, transformer blocks are selectively introduced within each stage to capture long-range dependencies and global context. These blocks operate based on the model's configuration, ensuring a balance between local and global feature extraction. At each stage $i$, a feature map $\mathbf{F}_i$ is generated, which is subsequently downsampled to reduce the spatial resolution. This operation is performed between stages to progressively capture both fine-grained details and high-level semantics:
\begin{equation}
    \mathbf{F}_i = \text{Downsample}(\mathbf{F}_{i-1}, \mathbf{x}_{down}^i).
\end{equation}

The final output of the encoder is a multi-scale representation $\mathbf{F} = \{ \mathbf{F}_1, \mathbf{F}_2, \mathbf{F}_3 \}$, consisting of hierarchical feature maps that preserve both local and global contextual information. These features are then passed to the decoder for further processing, enabling effective reconstruction and depth estimation.

\begin{algorithm}[ht]
\small
\caption{Encoder}
\label{alg:Encoder}
\LinesNumbered
\KwIn{Input image $x$, model configuration $config$}
\KwOut{Encoded features $F$}
\SetKwFunction{FForward}{Forward}
\SetKwFunction{FForwardFeatures}{ForwardFeatures}
\SetKwFunction{FDownsample}{Downsample}
\SetKwFunction{FStageProcess}{StageProcess}
\SetKwProg{Fn}{Function}{:}{}

\Fn{\FForward{$x$}}{
    $F \gets$ \FForwardFeatures{$x$}\;
    \Return $F$\;
}

\Fn{\FForwardFeatures{$x$}}{
    $x \gets (x - \mu) / \sigma $\;
    $x\_down \gets$ [AvgPool($x$, $i$) for $i$ in range(1, 5)]\;
    $features \gets []$\;
    $x \gets$ Stem($x$, $x\_down[0]$)\;
    \For{$i \gets 0$ \KwTo $2$}{
        $x \gets$ \FStageProcess{$x$, $i$}\;
        $features$.append($x$)\;
        \If{$i < 2$}{
            $x \gets$ \FDownsample{$x$, $x\_down[i+1]$, $features$}\;
        }
    }
    \Return $features$\;
}

\Fn{\FStageProcess{$x$, $i$}}{
    \For{$j \gets 0$ \KwTo $depth[i] - 1$}{
        \If{$j > depth[i] - transformer\_block[i] - 1$}{
            $x \gets$ TransformerBlock($x$, $config$)\;
        }
        \Else{
            $x \gets$ ConvBlock($x$, $dilation[i][j]$)\;
        }
    }
    \Return $x$\;
}

\Fn{\FDownsample{$x$, $x\_down$, $features$}}{
    $tmp\_x \gets$ Concatenate($features$, $x\_down$)\;
    \Return Conv($tmp\_x$, $dims[i+1]$, stride=2)\;
}

\Fn{Stem{$x$, $x\_down$}}{
    $x \gets$ Conv($x$, $dims[0]$, stride=2)\;
    $x \gets$ Conv($x$, $dims[0]$, stride=1)\;
    $x \gets$ Conv($x$, $dims[0]$, stride=1)\;
    \Return Conv(Concatenate($x$, $x\_down$), $dims[0]$, stride=2)\;
}

\Fn{TransformerBlock{$x$, $config$}}{
    \If{$transformer\_block\_type == $ 'TFB'}{
        \Return TFB($x$, $config$)\;
    }
    \Else{
        \Return $x$\;
    }
}
\end{algorithm}


\subsubsection{Decoder Architecture Design}

The decoder is responsible for reconstructing the output, specifically depth maps and surface normals, from the encoded features $\mathbf{F}$. The decoding process, as outlined in Algorithm~\ref{alg:Decoder}, begins with a series of upsampling operations to progressively restore the spatial resolution of the feature maps. For each stage $i$, the decoder employs an upsampling convolution, denoted as \textit{Upconv}, to increase the resolution of the feature maps, expressed as:
\begin{equation}
    \mathbf{F}_{up}^{i} = \text{Upconv}(\mathbf{F}_{enc}^{i}).
\end{equation}

At each stage, skip connections are utilised to concatenate the feature maps from the encoder $\mathbf{F}_{enc}$ with the upsampled feature maps in the decoder $\mathbf{F}_{up}$. This strategy allows the model to merge high-level semantic information from the encoder with fine-grained details, enabling better reconstruction of complex spatial structures. The concatenation is defined as:
\begin{equation}
    \mathbf{F}_{dec}^{i} = \text{Concatenate}(\mathbf{F}_{up}^{i}, \mathbf{F}_{enc}^{i-1}).
\end{equation}

For specific scales $i \in S$, the decoder produces depth and surface normal predictions. The depth predictions, denoted by $\mathbf{D}_i$, are generated using a \textit{Dispconv} operation, followed by a sigmoid activation to ensure the depth values are bounded within a valid range:
\begin{equation}
    \mathbf{D}_i = \sigma\left(\text{Dispconv}\left(\mathbf{F}_{dec}^{i}\right)\right),
\end{equation}
where $\sigma$ represents the sigmoid function.

Surface normal predictions, $\mathbf{N}_i$, are obtained via a \textit{Normconv} operation. The predicted normals are then normalised to form unit vectors, ensuring physically plausible surface orientations. This is achieved by normalising the vector components $N_x$, $N_y$, and $N_z$:
\begin{equation}
    \mathbf{N}_i = \frac{\left(N_x, N_y, N_z\right)}{\sqrt{N_x^2 + N_y^2 + N_z^2} + \epsilon},
\end{equation}
where $\epsilon$ is a small constant added for numerical stability during normalisation.

By leveraging these operations, the decoder efficiently reconstructs the target depth maps and surface normals, maintaining a balance between accuracy and computational efficiency. The outputs $\mathbf{D}$ and $\mathbf{N}$ at different scales are passed to the subsequent modules for further refinement or application.


\begin{algorithm}[ht]
\small
\caption{Decoder}
\label{alg:Decoder}
\LinesNumbered
\KwIn{Encoder features $F$, scales $S$, use\_skips $U$}
\KwOut{Depth maps $D$, Normal maps $N$}
\SetKwFunction{FForward}{Forward}
\SetKwFunction{FUpconv}{Upconv}
\SetKwFunction{FDispconv}{Dispconv}
\SetKwFunction{FNormconv}{Normconv}
\SetKwFunction{FNormalize}{Normalize}
\SetKwProg{Fn}{Function}{:}{}

\Fn{\FForward{$F$, $S$, $U$}}{
    $x \gets F[-1]$\;
    outputs $\gets \{\}$\;
    \For{$i \gets 2$ \KwTo $0$ \textbf{step} $-1$}{
        $x \gets$ \FUpconv{$x$, $i$}\;
        \If{$U$ \textbf{and} $i > 0$}{
            $x \gets$ Concatenate($x$, $F[i-1]$)\;
        }
        $x \gets$ \FUpconv{$x$, $i$}\;
        \If{$i \in S$}{
            $disp \gets$ \FDispconv{$x$, $i$}\;
            $normal \gets$ \FNormconv{$x$, $i$}\;
            outputs[$i$] $\gets \{disp, normal\}$\;
        }
    }
    \Return outputs\;
}

\Fn{\FUpconv{$x$, $i$}}{
    \Return ConvBlock($x$, $channels\_out = num\_ch\_dec[i]$)\;
}

\Fn{\FDispconv{$x$, $i$}}{
    $disp \gets$ Conv3x3($x$, $channels\_out = 1$)\;
    $disp \gets$ Upsample($disp$)\;
    \Return Sigmoid($disp$)\;
}

\Fn{\FNormconv{$x$, $i$}}{
    $normal \gets$ Conv3x3($x$, $channels\_out = 3$)\;
    $normal \gets$ Upsample($normal$)\;
    \Return \FNormalize($normal$)\;
}

\Fn{\FNormalize($normal$)}{
    $norm\_x, norm\_y, norm\_z \gets$ SplitChannels($normal$)\;
    $norm \gets \sqrt{norm\_x^2 + norm\_y^2 + norm\_z^2} + \epsilon$\;
    \Return Concatenate($norm\_x / norm$, $norm\_y / norm$, $norm\_z / norm$)\;
}

\end{algorithm}


\subsubsection{Loss Function}

Addressing the inherent challenges in depth estimation, our approach focuses on predicting depth.  We use a set of scale- and shift-invariant loss functions \cite{midas}, ensuring that model predictions remain consistent across different scales and shifts in the data, enhancing the robustness of the training process.


Let $M$ denote the number of pixels in an image with valid ground truth, and let $\btheta$ represent the parameters of the prediction model. We define the predicted depth as $\dd = \dd(\btheta) \in \Re^M$ and the ground truth depth as $\dd^* \in \Re^M$. Individual pixel values are denoted by subscripts.

\paragraph{\textbf{Scale- and Shift-Invariant Loss}:}
To compute the loss for a single sample, we apply a scale- and shift-invariant loss, formulated as:

\begin{equation}
\lL_{ssi}(\hat \dd, \hat \dd^*) = \frac{1}{2M} \sum_{i=1}^M \rho \left( \hat \dd_i - \hat \dd_i^* \right),
\label{eq:ssi_loss}
\end{equation}

where $\hat \dd$ and $\hat \dd^*$ are the scaled and shifted versions of the predicted and ground truth disparities, respectively. The function $\rho$ defines the type of loss applied, allowing flexibility in the choice of error metric.

\paragraph{\textbf{Gradient Matching Term}:}
To refine the depth prediction, we employ a multi-scale, scale-invariant gradient matching term, inspired by \cite{midas}, to promote sharp discontinuities that align with the ground truth. This term is defined as:

\begin{equation}
  \lL_{reg}(\hat \dd, \hat \dd^*) = \frac{1}{M} \sum_{k=1}^K \sum_{i=1}^{M} \left( |\nabla_x R_i^k| + |\nabla_y R_i^k| \right),
  \label{eq:grad_loss}
\end{equation}

where $R_i = \hat \dd_i - \hat \dd_i^*$ represents the difference between predicted and ground truth depth values, and $R^k$ refers to the difference at scale $k$. We use $K=4$ scales, progressively reducing the resolution by half at each level.

\paragraph{\textbf{Final Loss}:}
The overall loss for a training set $l$ is a combination of the scale- and shift-invariant loss (Equation~\ref{eq:ssi_loss}) and the gradient matching term (Equation~\ref{eq:grad_loss}):

\begin{equation}
  \lL_l = \frac{1}{N_l}\sum_{n=1}^{N_l} \left( \lL_{ssi}\big(\hat \dd^n, (\hat \dd^*)^n\big) + \alpha \; \lL_{reg}\big(\hat \dd^n, (\hat \dd^*)^n\big) \right),
  \label{eq:total_loss}
\end{equation}

where $N_l$ is the size of the training set, and the weighting parameter $\alpha$ is set to $0.5$.

This carefully designed loss structure ensures both accurate depth prediction and the precise alignment of salient features, such as edges, between the predicted and ground truth disparities. This approach ensures that the model produces high-quality, affine-invariant inverse depth maps while effectively mitigating the impact of noisy pseudo-labels during the training process.  We apply the same loss formulation to surface normals, ensuring consistency and accuracy across varying scales and shifts, ultimately contributing to the overall robustness of the model.




\section{Experiments}  \label{sec:experiments}

This section evaluates the performance of our proposed method for underwater monocular depth and surface normals estimation. We benchmark our approach against state-of-the-art methods, focusing on key metrics such as accuracy, inference speed, and computational cost, to highlight the model's efficacy in resource-constrained environments. Extensive tests on real-world underwater datasets demonstrate the accuracy, computational efficiency, and generalisation capabilities of our lightweight hybrid CNN-Transformer model, particularly for real-time applications on edge devices.

\subsection{Implementation Details}
The proposed method is implemented in PyTorch and trained on a Linux platform using a single NVIDIA GeForce RTX 4090 GPU, with a batch size of 22. The Adam optimizer \citep{kingma2014adam} is employed, using $\beta_1 = 0.9$, $\beta_2 = 0.999$, and $\epsilon = 1.0 \times 10^{-8}$. The initial learning rate is set to $lr_{initial} = 1 \times 10^{-4}$, which linearly decays to $lr_{final} = 1 \times 10^{-6}$ over the final $100$ epochs. The model was trained on RGB-D image pairs, covering diverse underwater scenes, for a total of $400$ epochs. The complete training process takes approximately $48$ hours on the specified hardware.

\subsection{Datasets} 

This research used a variety of publicly available underwater image datasets, as detailed in Table~\ref{tab:datssets} to train and evaluate the performance of our proposed method. These datasets represent diverse underwater environments with variations in water clarity, lighting conditions, and object types. The data sets were divided into different training and testing sets. 
Unlike previous methods, our approach did \textbf{ not} rely on ground-truth depth maps or surface normals during training, highlighting the strength of our pseudo-labelling technique.
Using these data sets, we ensured a robust evaluation of our method for both depth and surface normal estimation in challenging underwater scenarios. 

\begin{table}[ht]
    \centering
    \caption{Dataset Overview}
    \label{tab:datssets}
    \begin{tabular}{|l|l|c|l|}
        \hline
        \textbf{Dataset} & \textbf{Type} & \textbf{Samples} & \textbf{Key Characteristics} \\ \hline
        USOD10K \cite{USOD10K} & Train/Test & 9229/1026 & $640\times480$ RGB images with depth maps from 12 scenes \\ \hline
        DRUVA \cite{DRUVA} & Train/Test & 2000 & Video sequences of submerged rocks, shallow waters \\ \hline
        DeepFish  \cite{DeepFish} & Train/Test & 5000 & Underwater images from tropical Australian waters \\ \hline
        Atlantis \cite{atlantis} & Test & 3200 & $672\times512$ images, varied underwater environments \\ \hline
        Sea-Thru \cite{Sea-Thru} & Test & 1100 & Depth-varied underwater images for depth estimation \\ \hline
        SQUID \cite{SQUID}& Test & 114 & Stereo image dataset with distance maps \\ \hline
        iBims-1 \cite{IBims} & Train/Test & 108 & RGB-D dataset for surface normal estimation benchmark \\ \hline
    \end{tabular}
\end{table}

\subsection{Evaluation Metrics}

We employed widely used evaluation metrics to rigorously assess the performance of our model. These metrics are commonly used for both depth estimation and normal estimation tasks, allowing for a standardised comparison with existing methods \cite{wang2024umono, udepth, litemono}.

\noindent
\textbf{Depth Estimation Metrics:}  
Following established practices in the field \cite{yang2024depth, bhat2023zoedepth, zavadski2024primedepth}, we assessed depth estimation performance using the following metrics:

\begin{itemize}
    \item \textbf{Absolute Relative Error (Abs Rel):} 
    $
    \text{AbsRel} = \frac{1}{N} \sum_{i=1}^{N} \frac{|d_i - d_i^*|}{d_i^*}
    $
    
    \item \textbf{Squared Relative Error (Sq Rel):}
    $
    \text{SqRel} = \frac{1}{N} \sum_{i=1}^{N} \frac{(d_i - d_i^*)^2}{d_i^*}
    $

    \item \textbf{Root Mean Squared Error (RMSE):}
    $
    \text{RMSE} = \sqrt{\frac{1}{N} \sum_{i=1}^{N} (d_i - d_i^*)^2}
    $

    \item \textbf{Logarithmic Error (log10):}
    $
    \log_{10} = \frac{1}{N} \sum_{i=1}^{N} |\log_{10}(d_i) - \log_{10}(d_i^*)|
    $

    \item \textbf{Accuracy:} The percentage of predicted depths $d_i$ such that:
    $
    \max \left( \frac{d_i}{d_i^*}, \frac{d_i^*}{d_i} \right) < \delta
    $
    Where $\delta$ is typically set to $\tau = 1.25$, $1.25^2$, or $1.25^3$, depending on the evaluation threshold.
\end{itemize}

Here, $d_i$ is the predicted depth, $d_i^*$ is the ground truth depth, and $N$ is the total number of pixels in the depth map.

\noindent
\textbf{Surface Normals Estimation Metrics:}  
The evaluation of surface normals estimation relied on several error metrics designed to quantify angular error:

\begin{itemize}
    \item \textbf{Mean Angular Error:} The mean angular error between predicted normals $n_i$ and ground truth normals $n_i^*$:
    $
    \text{Mean} = \frac{1}{N} \sum_{i=1}^{N} \cos^{-1}(n_i \cdot n_i^*)
    $

    \item \textbf{Median Angular Error:} The median of the angular errors:
    $
    \text{Median} = \text{Median} \left( \cos^{-1}(n_i \cdot n_i^*) \right)
    $

    \item \textbf{Root Mean Squared Error (RMS Normal):} The root mean squared of the angular errors:
    $
    \text{RMS Normal} = \sqrt{\frac{1}{N} \sum_{i=1}^{N} \left( \cos^{-1}(n_i \cdot n_i^*) \right)^2}
    $

    \item \textbf{Accuracy:} The percentage of predicted normals where the angular error is less than a threshold, typically set to $11.25^\circ$, $22.5^\circ$, or $30^\circ$.
\end{itemize}

\subsection{Quantitative Comparison}

To fairly evaluate the performance of our proposed model, AquaMDS, we performed a quantitative comparison against several state-of-the-art methods of similar size for both depth estimation and surface normals estimation, ensuring a fair comparison without including significantly larger models.  All baseline model results were obtained from the published articles of the respective authors to ensure a fair and accurate comparison.

\textbf{Depth Estimation}:  
We evaluated the depth estimation capabilities of AquaMDS on three diverse underwater datasets: USOD 10K \cite{USOD10K}, Atlantis \cite{atlantis} , and Sea-thru \cite{Sea-Thru}.  These datasets were chosen to assess the model's performance across various underwater scenarios and its ability to generalise to unseen data.  

\noindent
\textbf{USOD 10K:}  As shown in table \ref{table:comparison}, AquaMDS significantly outperformed all other methods. It achieved the highest accuracy scores across all three thresholds ($\delta_1$, $\delta_2$, and $\delta_3$) while simultaneously achieving the lowest error rates (Abs Rel, Sq Rel, RMSE, and log10).  

\noindent
\textbf{Atlantis:} Similar to its performance on USOD 10K, AquaMDS demonstrated superior performance on the Atlantis dataset, table \ref{table:comparison2}, surpassing other methods in both accuracy and error metrics. 

\noindent
\textbf{Sea-thru:} On the Sea-thru (D3) dataset, AquaMDS again achieved the best results across all evaluation metrics, further validating its effectiveness in accurate depth estimation for diverse underwater scenes. Table Table \ref{tab:seathru} provides a detailed breakdown of the results, with AquaMDS consistently achieving the lowest error rates.

\textbf{Surface Normals Estimation:}  
Evaluating surface normal estimation in underwater environments presents a unique challenge due to the lack of an established benchmark or dataset specifically designed for this task.  To address this limitation, we trained and evaluated AquaMDS on the iBims-1 dataset \cite{IBims}. Although iBims-1 was not created for underwater images, it serves as a robust benchmark for assessing surface normal estimation techniques.  
As presented in Table \ref{table:ibims1_normal}, AquaMDS achieved competitive results on the iBims-1 benchmark. It performed strongly across key metrics like mean and median errors, as well as accuracy thresholds at various angles.  

\begin{table*}[ht]
\centering
\caption{Quantitative comparison results for depth estimation of our proposed method and compared methods on USOD 10K \cite{USOD10K} dataset. For $\delta_1$, $\delta_2$, and $\delta_3$, ($\uparrow$) means higher is better. For other metrics, ($\downarrow$) means that lower is better. }
\label{table:comparison}
\begin{tabular}{l|ccccccc}
\hline
\textbf{Method} & $\delta_1$ $\uparrow$& $\delta_2$ $\uparrow$& $\delta_3$ $\uparrow$& \textbf{Abs Rel} $\downarrow$& \textbf{Sq Rel} $\downarrow$& \textbf{RMSE} $\downarrow$& \textbf{log10} $\downarrow$\\
\hline
GDCP \cite{gdcp} & 0.2100 & 0.3878 & 0.5376 & 1.6465 & 0.6501 & 0.2963 & 0.3444 \\
UW-Net \cite{uw-net} & 0.2276 & 0.3846 & 0.5101 & 2.5464 & 1.4091 & 0.3894 & 0.3753 \\
BTS \cite{bts} & 0.4242 & 0.6709 & 0.8010 & 0.6634 & 0.1154 & 0.1476 & 0.1876 \\
LapDepth \cite{lapdepth} & 0.4664 & 0.6993 & 0.8238 & 0.5885 & 0.1005 & 0.1327 & 0.1709 \\
LiteMono \cite{litemono} & 0.4629 & 0.6897 & 0.8111 & 0.5406 & 0.0940 & 0.1344 & 0.1811 \\
TransDepth \cite{transdepth} & 0.4636 & 0.7085 & 0.8267 & 0.5405 & 0.0910 & 0.1324 & 0.1780 \\
UDepth \cite{udepth} & 0.4508 & 0.6723 & 0.7935 & 0.6810 & 0.1230 & 0.1430 & 0.1880 \\
UMono  \cite{wang2024umono} & {0.5237} & {0.7471} & {0.8519} & {0.4956} & {0.0778} & {0.1182} & {0.1526} \\
AquaMDS (Ours) & \textbf{0.7546} & \textbf{0.8065} & \textbf{0.9224} & \textbf{0.3511} & \textbf{0.0457} & \textbf{0.0952} & \textbf{0.0943} \\
\hline
\end{tabular}
\end{table*}

\begin{table*}[ht]
\centering
\caption{Quantitative comparison results for depth estimation of our proposed method and compared methods on Atlantis  \cite{atlantis} dataset. For $\delta_1$, $\delta_2$, and $\delta_3$, ($\uparrow$) means higher is better. For other metrics, ($\downarrow$) means that lower is better. }
\label{table:comparison2}
\begin{tabular}{l|ccccccc}
\hline
\textbf{Method} & $\delta_1$ $\uparrow$& $\delta_2$ $\uparrow$& $\delta_3$ $\uparrow$& \textbf{Abs Rel} $\downarrow$& \textbf{Sq Rel} $\downarrow$& \textbf{RMSE} $\downarrow$& \textbf{log10} $\downarrow$\\
\hline
GDCP \cite{gdcp} & 0.1636 & 0.3349 & 0.4762 & 3.3197 & 2.7447 & 0.4540 & 0.4108 \\
UW-Net \cite{uw-net} & 0.1489 & 0.2887 & 0.4302 & 1.6701 & 0.9291 & 0.4504 & 0.3610 \\
BTS \cite{bts} & 0.3753 & 0.6017 & 0.7366 & 1.1198 & 0.2188 & 0.1550 & 0.2260 \\
LapDepth \cite{lapdepth} & 0.4062 & 0.6221 & 0.7501 & 1.0856 & 0.2041 & 0.1443 & 0.2144 \\
LiteMono \cite{litemono} & 0.4047 & 0.6298 & 0.7545 & 0.8493 & 0.1573 & 0.1456 & 0.2117 \\
TransDepth \cite{transdepth} & 0.4078 & 0.6254 & 0.7602 & 0.8851 & 0.1580 & 0.1443 & 0.2060 \\
UDepth \cite{udepth} & 0.3616 & 0.5632 & 0.6975 & 1.2664 & 0.3231 & 0.1799 & 0.2451 \\
UMono  \cite{wang2024umono} & {0.4096} & {0.6354} & {0.7694} & {0.8449} & {0.1450} & {0.1422} & {0.2017} \\
AquaMDS (Ours) & \textbf{0.6324} & \textbf{0.8141} & \textbf{0.9182} & \textbf{0.6241} & \textbf{0.0932} & \textbf{0.0944} & \textbf{0.1054} \\
\hline
\end{tabular}
\end{table*}

\begin{table}[ht]
  \centering
  \setlength{\tabcolsep}{1.5mm}{
    \caption{Quantitative performance comparison on Sea-thru (D3) dataset \cite{Sea-Thru}. Lower scores represent better performance for all metrics.}
    \label{tab:seathru}
    \begin{tabular}{l|cccc}
      \toprule
      Model & Abs Rel & Sq Rel & RMSE & log10 \\
      \midrule
      AdaBins (RGB) \cite{adabins} & 1.314 & 0.663 & 0.441 & 0.356 \\
      AdaBins (RMI) \cite{adabins} & 1.258 & 0.617 & 0.426 & 0.346 \\
      DenseDepth (RGB) \cite{dense_depth} & 1.073 & 0.448 & 0.366 & 0.312 \\
      DenseDepth (RMI) \cite{dense_depth} & 1.092 & 0.470 & 0.370 & 0.312 \\
      UDepth (RGB) \cite{udepth} & 1.304 & 0.653 & 0.440 & 0.355 \\
      UDepth (RMI) \cite{udepth} & {1.153} & {0.514} & 0.388 & 0.321 \\
      AquaMDS (Ours)& \textbf{1.014} & \textbf{0.414} & \textbf{0.317} & \textbf{0.294} \\
      \bottomrule
    \end{tabular}
  }
\end{table}

\begin{table*}[ht]
\centering
\caption{ibims-1 datset \cite{IBims} Normal Benchmark comparison. Higher is better for 11.25\textdegree , 22.5\textdegree , and 30\textdegree  ($\uparrow$), while lower is better for mean, median, and RMS normal error ($\downarrow$).}
\label{table:ibims1_normal}
\begin{tabular}{l|ccc|ccc}
\hline
\textbf{Method} & \textbf{11.25\textdegree } $\uparrow$ & \textbf{22.5\textdegree } $\uparrow$ & \textbf{30\textdegree } $\uparrow$ & \textbf{Mean} $\downarrow$ & \textbf{Median} $\downarrow$ & \textbf{RMS Normal} $\downarrow$ \\
\hline
VNL \cite{vnl} & 0.179 & 0.386 & 0.494 & 39.8 & 30.4 & 51.0 \\
BTS \cite{bts} & 0.130 & 0.295 & 0.400 & 44.0 & 37.8 & 53.5 \\
Adabins \cite{adabins} & 0.180 & 0.387 & 0.506 & 37.1 & 29.6 & 46.9 \\
IronDepth \cite{irondepth} & 0.431 & 0.639 & 0.716 & 25.3 & 14.2 & 37.4 \\
Omnidata \cite{omnidata} & 0.647 & 0.734 & 0.768 & 20.8 & 7.7 & 35.1 \\
Metric3D v2 \cite{yin2023metric3d} & \textbf{0.697} & \textbf{0.762} & \textbf{0.788} & \textbf{19.6} & \textbf{5.7} & \textbf{35.2} \\
AquaMDS (Ours) & 0.634 & 0.738 & 0.775 & 20.4 & 6.7 & 36.9 \\
\hline
\end{tabular}
\end{table*}

\begin{table}[htbp]
\centering
\caption{Quantitative Comparison Results of Ablation Study on the Benefit of Conv and Transformer Block. The value in \textbf{bold}  means the best results.}
\label{table:ablation_study_1}
\begin{tabular}{|l|c|c|c|c|c|}
\hline
\textbf{Variant} & $\delta_3$ $\uparrow$& \textbf{Abs Rel} $\downarrow$& \textbf{Sq Rel} $\downarrow$& \textbf{RMSE} $\downarrow$& \textbf{log10} $\downarrow$\\ 
\hline
w/ Conv         & 0.9120 & 0.3602 & 0.0501 & 0.1034 & 0.1078 \\ 
w/ Transformer & 0.9153 & 0.3553 & 0.0483 & 0.0981 & 0.0967 \\ 
w/o Feature Fusion       & 0.9207 & 0.3543 & 0.0461 & 0.0964 & 0.0950 \\ 
Full model           & \textbf{0.9224} & \textbf{0.3511} & \textbf{0.0457} & \textbf{0.0952} & \textbf{0.0943} \\
\hline
\end{tabular}
\end{table}

\begin{table}[htbp]
\centering
\caption{Quantitative Comparison Results of Ablation Study on the Benefit of dilated convolutions, pooled concatenations and cross-stage connections. The value in \textbf{bold} means the best results.}
\label{table:ablation_study_2}
\begin{tabular}{|l|c|c|c|c|c|}
\hline
\textbf{Variant} & $\delta_3$ $\uparrow$& \textbf{Abs Rel} $\downarrow$& \textbf{Sq Rel} $\downarrow$& \textbf{RMSE} $\downarrow$& \textbf{log10} $\downarrow$\\ 
\hline
w/o dilated convolutions        & 0.9202 & 0.3520 & 0.0465 & 0.0965 & 0.0954 \\ 
w/o pooled concatenations & 0.9211 & 0.3515 & 0.0461 & 0.0963 & 0.0950 \\ 
w/o cross-stage connections       & 0.9215 & 0.3513 & 0.0459 & 0.0957 & 0.0948 \\ 
Full model           & \textbf{0.9224} & \textbf{0.3511} & \textbf{0.0457} & \textbf{0.0952} & \textbf{0.0943} \\
\hline
\end{tabular}
\end{table}

\begin{figure}[ht]
    \centering
    \includegraphics[width=0.98\linewidth]{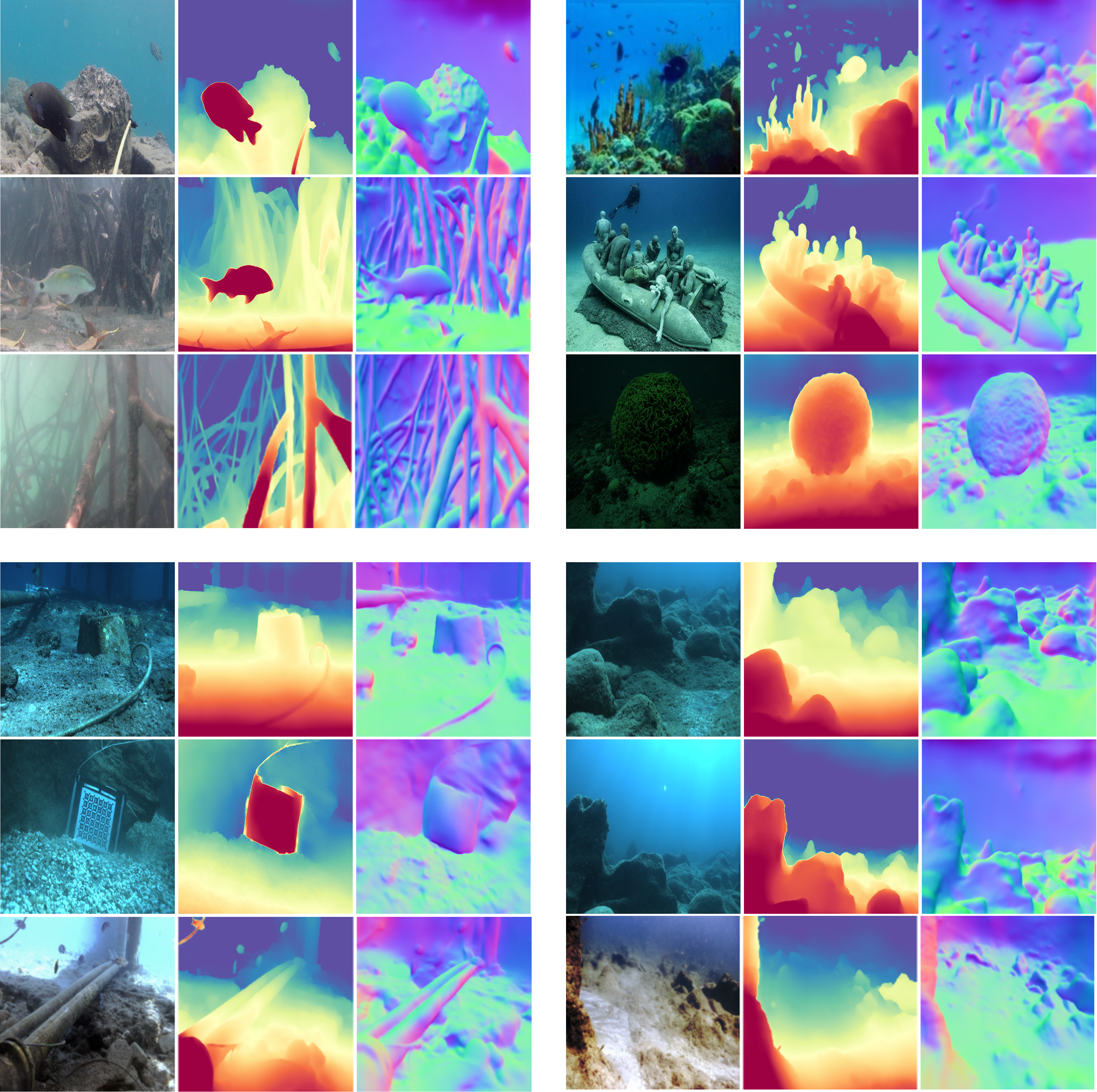}
    \caption{Qualitative comparison of our model's depth and surface normal predictions, trained exclusively on pseudo-labelled real-world images.  From top left: DeepFish \cite{DeepFish}, USOD10K \cite{USOD10K}, SQUID \cite{SQUID}, and Sea-Thru \cite{Sea-Thru}.  These results demonstrate the model's robustness and ability to capture fine-grained details. 
}
    \label{l_fig_5}
\end{figure}

\begin{figure}[ht]
    \centering
    \includegraphics[width=0.98\linewidth]{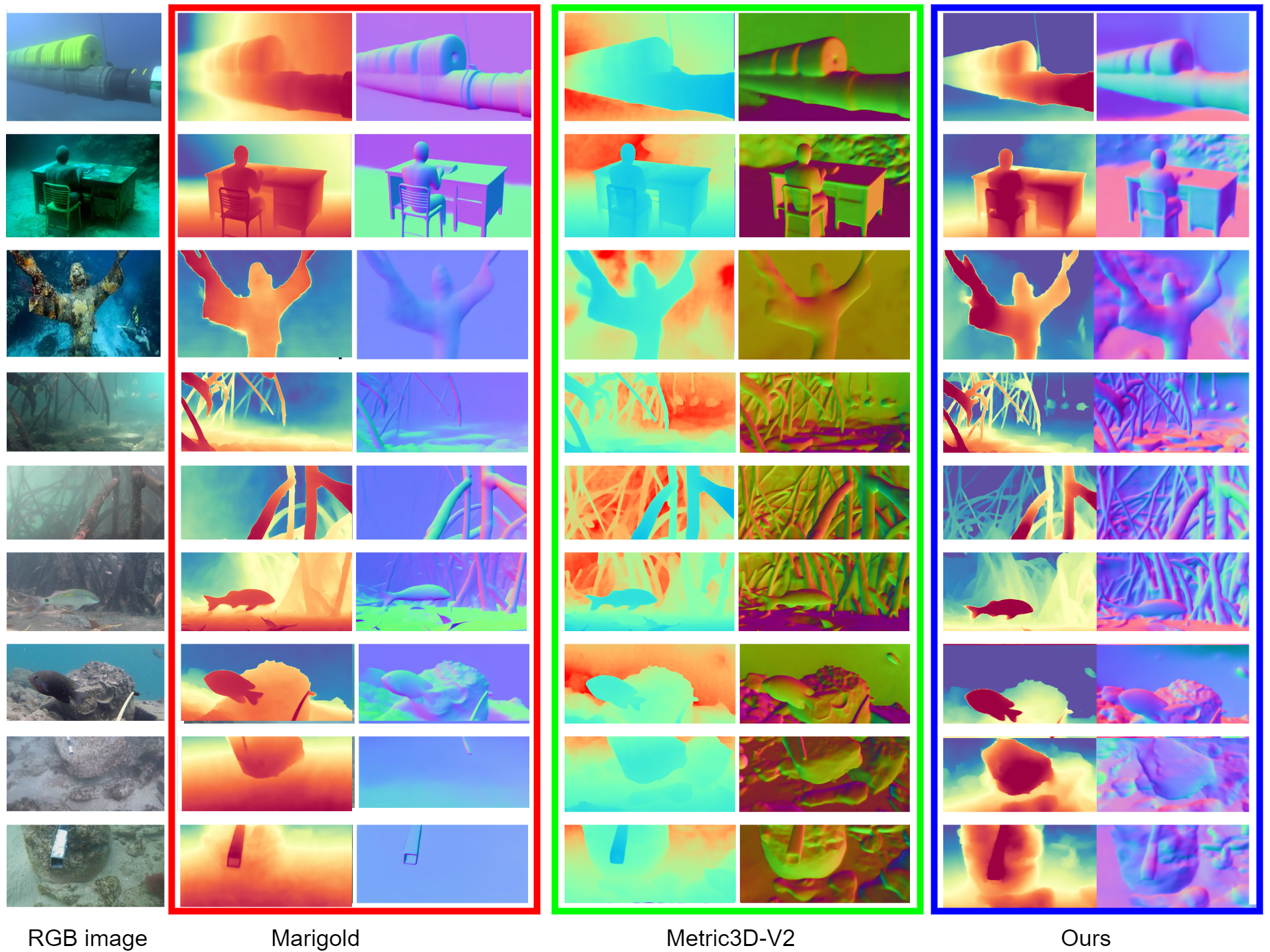}
    \caption{Comparison between Marigold \cite{ke2024repurposing}, Metric3D-V2 \cite{yin2023metric3d}, and our model for depth and surface normals in open-world images. These predictions are from their public HuggingFace best model \cite{face2021ai}. \textit{The figure may be better viewed online where you can zoom in for more detail.}}
    \label{l_fig_6}
\end{figure}

\subsection{Qualitative Comparison}

To further evaluate our proposed method, we conducted a qualitative comparison against existing models, focusing on both depth estimation and surface normals estimation. This qualitative assessment complements the quantitative results, providing visual insights into the model's performance. 

Figure \ref{l_fig_5} showcases the outputs of our model trained exclusively on pseudo-labeled real images.  The results demonstrate the model's robustness in capturing intricate details in both depth and surface normals, particularly in visually complex underwater environments. 

In Figure \ref{l_fig_6}, we present a comparative analysis between our model (AquaMDS), Marigold \cite{ke2024repurposing}, and Metric3D-V2 \cite{yin2023metric3d}, using open-world images. All models were evaluated using their best publicly available versions from the HuggingFace model hub \cite{face2021ai}. The results consistently demonstrate that our method delivers more refined and accurate predictions for both depth and surface normals, especially in challenging scenarios where existing models, such as Marigold and Metric3D-V2, tend to exhibit limitations. Our approach demonstrates a noticeable improvement in sharpness and structure preservation, further illustrating its effectiveness in handling real-world imagery.  


\subsection{Ablation Study}

To thoroughly investigate the contribution of each component in our proposed model, we conducted an extensive ablation study. Our analysis focused on evaluating the performance impact of various architectural elements, specifically the convolutional and transformer blocks, dilated convolutions, pooled concatenations, and cross-stage connections.

\begin{enumerate}
    \item \textbf{Convolutional and Transformer Blocks}: The study revealed that both the convolutional ($\text{Conv}$) and transformer ($\text{Transformer}$) blocks are critical to the model's overall performance. Removing either block resulted in a significant decrease in accuracy, underscoring the importance of the synergy between these two components. This observation is supported by the quantitative results presented in Table~\ref{table:ablation_study_1}, where the full model, which includes both convolutional and transformer blocks, achieves the best performance.
    
    \item \textbf{Dilated Convolutions, Pooled Concatenations, and Cross-Stage Connections}: Similarly, the removal of any of the following components-dilated convolutions, pooled concatenations, or cross-stage connections-led to a degradation in the model's performance. This suggests that each of these architectural elements plays a pivotal role in enhancing depth and surface normal estimation accuracy. The quantitative comparison results for these components are summarised in Table~\ref{table:ablation_study_2}, where the full model consistently outperforms the variants lacking these features.
\end{enumerate}


\subsection{Computational Efficiency}

Beyond accuracy, we also analysed AquaMDS's computational efficiency, a crucial factor for real-time applications and deployment on resource-constrained devices. As illustrated in Table \ref{tab:FPS}, AquaMDS demonstrated a notable balance between accuracy and efficiency. It boasts a significantly lower parameter count and requires less memory compared to more complex models like TransDepth.  Furthermore, AquaMDS achieves a high frame rate (FPS), enabling real-time performance with minimal computational resources.

\begin{table}[htbp]
\centering
\caption{Comparison Results for Complexity and FPS of the proposed AquaMDS and compared methods. On a Linux platform using a single NVIDIA GeForce RTX 4090 GPU}
\label{tab:FPS}
\begin{tabular}{|l|c|c|c|c|}
\hline
\textbf{Method} & \textbf{Params (M)} & \textbf{Memory (GB)} & \textbf{FLOPs (G)} & \textbf{FPS} \\ \hline
BTS  \cite{bts}      & 16.34  & 62.65  & 246.80  & 26  \\ 
LapDepth  \cite{lapdepth} & 73.57  & 281.62 & 364.57  & 20   \\ 
LiteMono \cite{litemono}  & 8.78   & 33.60  & 112.12  & 55  \\ 
TransDepth \cite{transdepth} & 247.40 & 944.13 & 1187.67 & 6   \\ 
UDepth  \cite{udepth}   & 15.65  & 59.78  & 77.16   & 66  \\ 
UMono \cite{wang2024umono}     & 28.11  & 107.61 & 148.91  & 30  \\ 
AquaMDS (Ours)& 7.58   & 30.60  & 98.10  & 63  \\ 
\hline
\end{tabular}
\end{table}



\section{Discussion} \label{sec:discussion}

This research introduces a novel approach to underwater monocular depth and surface normals estimation, addressing the limitations of existing methods while prioritising efficiency and practical applicability. By combining a hybrid CNN-Transformer architecture with a novel pseudo-labelling technique, this work achieves state-of-the-art performance with significantly reduced computational demands.

\paragraph{Balancing Local and Global Features:}
The success of our proposed AquaMDS model stems from its hybrid architecture, which effectively integrates the strengths of CNNs and Transformers. CNNs excel at extracting local features and details, while Transformers are adept at capturing long-range dependencies and global context. This combination is particularly valuable in underwater environments, where variations in lighting, turbidity, and object textures complicate accurate depth and surface normal estimation.

\paragraph{Enhancing Generalisation with Pseudo-Labelling:}
Recognising the limitations of relying solely on real-world datasets, often plagued by noisy or missing labels, this research introduces a novel pseudo-labelling approach. This method leverages the strengths of existing pre-trained models to generate high-quality pseudo-labelled data from readily available unlabeled underwater images. To ensure the accuracy of these pseudo-labelled data, we developed DNESA (Depth Normal Evaluation and Selection Algorithm), which uses domain-specific metrics to evaluate and select the most reliable samples for training. This approach effectively bridges the gap between synthetic and real-world data, resulting in a more robust and generalisable model.

\paragraph{Practicality Through Efficiency:}
A key contribution of this work is its focus on computational efficiency without compromising accuracy. The proposed model achieves a notable reduction in parameters 90\% and a reduction in the computational cost of training 80\% compared to current methods. This efficiency translates to real-time inference capabilities (approximately 15.87ms per image), making the model highly suitable for deployment on resource-constrained devices commonly used in underwater robotics and autonomous vehicles. This advancement opens up new possibilities for real-time 3D perception in underwater environments.

\paragraph{Broader Implications for Marine Science:}
The implications of this research extend beyond the improved estimate of the depth and surface normals. The ability to generate accurate 3D information from a single image, efficiently and in real time, holds immense potential for various marine science applications, including:

\begin{itemize}
    \item \textbf{Detailed Seafloor Mapping:} High-resolution depth maps facilitate the creation of more accurate and detailed seafloor maps, essential for understanding underwater topography and ecosystems.
    \item \textbf{Coral Reef Health Assessment:} By providing detailed 3D models of coral reefs, this technology aids in monitoring their health, identifying areas of damage or disease, and informing conservation efforts.
    \item \textbf{Species Population Studies:} Accurate depth information can be used to estimate fish populations and track their movements, providing valuable data for fisheries management and ecological research.
    \item \textbf{Monitoring Underwater Infrastructure:} This technology assists in inspecting and monitoring the structural integrity of underwater infrastructure, such as pipelines and submerged structures.
\end{itemize}

The efficient and scalable nature of our proposed model makes it a valuable tool for both researchers and practitioners in marine science, offering a practical solution for a wide range of underwater applications.

\section{Conclusions}  \label{sec:conclusion}

This research presents a practical solution for the estimation of monocular depth and surface normals in underwater environments. By combining a hybrid CNN-Transformer architecture with an innovative pseudo-labelling technique, the proposed model achieves high accuracy while significantly reducing computational demands and reliance on large labelled datasets.
AquaMDS, our hybrid model, is designed for efficient depth and surface normal estimation, with DNESA improving the quality of pseudo-labelled data and enhancing generalisation. This approach bridges the gap between research and real-world application, enabling real-time 3D perception on resource-constrained devices.
Our model is particularly suited for underwater tasks such as seafloor mapping, coral reef monitoring, and species population studies. Future work will expand the model's capabilities for broader marine applications and improve its integration into real-world systems.




\section*{Data Availability}
\label{sec:Data_Availability}

The datasets are publicly available on their corresponding online repositories.

\section*{Acknowledgement}
This research was supported by the Australian Institute of Marine Science (AIMS) and James Cook University (JCU) through a Postdoctoral Research Fellowship under the AIMS@JCU program. The authors would like to thank AIMS and JCU for their continued support and resources provided during the course of this work.


\section*{Funding}
No funding was received for this research.



\section*{Additional Information}
\textbf{Competing interests} The authors declare no competing interests.\\


	
\printcredits

\bibliographystyle{cas-model2-names}

\bibliography{references}


 

%

\end{document}